\documentclass[10pt,twocolumn,letterpaper]{article}

\usepackage{iccv}
\usepackage{times}
\usepackage{epsfig}
\usepackage{graphicx}
\usepackage{amsmath}
\usepackage{amssymb}
\usepackage{booktabs}
\usepackage{enumerate} 
\usepackage{multirow}
\usepackage{colortbl}
\usepackage{subcaption}
\usepackage{wrapfig}
\newcommand{\eat}[1]{}
\usepackage[pagebackref=true,breaklinks=true,letterpaper=true,colorlinks,bookmarks=false]{hyperref}

\def\iccvPaperID{7404} 

\ificcvfinal\fi

\usepackage[capitalize]{cleveref}
\crefname{section}{Sec.}{Secs.}
\Crefname{section}{Section}{Sections}
\Crefname{table}{Table}{Tables}
\crefname{table}{Tab.}{Tabs.}

\begin{document}

\title{CIParsing: Unifying Causality Properties into Multiple Human Parsing}

\author{\bf Xiaojia Chen, \bf Xuanhan Wang, \bf Lianli Gao, \bf Beitao Chen, \bf Jingkuan Song, \bf HenTao Shen \\ \\
Center for Future Media, \\
University of Electronic Science and Technology of China}

\maketitle
\ificcvfinal\thispagestyle{empty}\fi
\begin{abstract}
Existing methods of multiple human parsing (MHP) apply statistical models to acquire underlying associations between images and labeled body parts.
However, acquired associations often contain many spurious correlations that degrade model generalization, leading statistical models to be vulnerable to visually contextual variations in images (e.g., unseen image styles/external interventions). 
To tackle this, we present a causality inspired parsing paradigm termed CIParsing, which follows fundamental causal principles involving two causal properties for human parsing (i.e., the causal diversity and the causal invariance). 
Specifically, we assume that an input image is constructed by a mix of causal factors (the characteristics of body parts) and non-causal factors (external contexts), where only the former ones cause the generation process of human parsing.
Since causal/non-causal factors are unobservable, a human parser in proposed CIParsing is required to construct latent representations of causal factors and learns to enforce representations to satisfy the causal properties. In this way, the human parser is able to rely on causal factors w.r.t relevant evidence rather than non-causal factors w.r.t spurious correlations, thus alleviating model degradation and yielding improved parsing ability. Notably, the CIParsing is designed in a plug-and-play fashion and can be integrated into any existing MHP models. 
Extensive experiments conducted on two widely used benchmarks demonstrate the effectiveness and generalizability of our method. 
Anonymous code and models are released\footnote{\url{https://anonymous.4open.science/r/COSParser}} for research purpose.
\end{abstract}
\section{Introduction}
\label{sec:intro}
Multiple human parsing, which is a fundamental task in computer vision, requires to segment each person in an image into semantically consistent regions belonging to the body parts. Optimally addressing this task would greatly support a wide range of downstream applications, such as texture transfer \cite{dp_transfer, WangKTN++} and fashion analysis \cite{dataset:fashionpedia,dataset:deepfashion2}. 

\begin{figure}[t]
\centering
\setlength{\abovecaptionskip}{3.pt}
\includegraphics[width=0.9\linewidth]{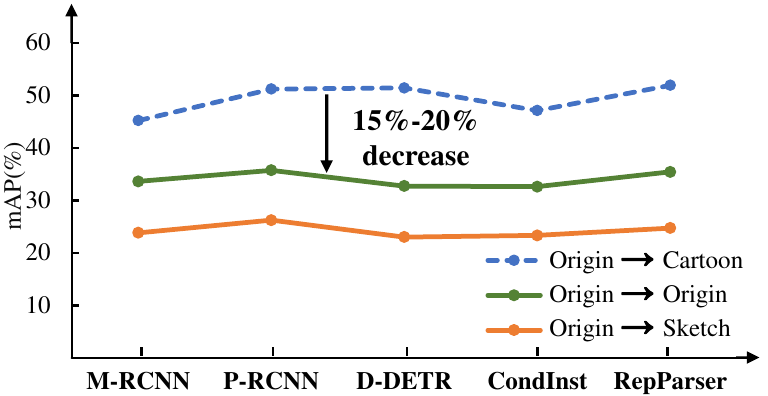}
\caption{The phenomenon of low-qualified generalizability: Existing parsing models suffer from a significant performance drop on images with unseen styles.}
\label{fig:domain_change}
\vspace{-1.5em}
\end{figure}
The key challenge in multiple human parsing is how to obtain instance-level body parts, involving two core sub-problems: (1) instance separation; and (2) body part segmentation. Mainstream methods address this challenge by adopting a two-stage strategy, including top-down and bottom-up approaches. In particular, the former approaches \cite{ins_seg:mask_rcnn,WangAMA++,densepose:parsingrcnn,parsing:semtree,parsing:mce2p,parsing:braidnet} often start with person detection responsible for addressing the first sub-problem, and then perform single-person part segmentation for tackling the second sub-problem. In contrast, the latter approaches \cite{seg_part:cihp,Parsing:Graphonomy,parsing:mhpwild,seg_part:mhp} first handle the second sub-problem by addressing part-level semantic segmentation, and then tackle the first sub-problem by instance-aware assembling. Although two-stage methods often obtain encouraging results, the use of isolated models separately responsible for sub-problems leads to heavily computational burden \cite{RepParser,Mao2021pose}. Furthermore, the predictive performance in the first stage severely limits the predictive performance in the second stage. To relieve these, recent attempts \cite{RepParser,obj_det:deformdetr,inst_condinst,obj_det:detr} focus on the one-stage paradigm, aiming to break down the dependence in the two-stage by a kernel-based framework. 

\begin{figure}[t]
\centering
\setlength{\abovecaptionskip}{3.pt}
\includegraphics[width=0.99\linewidth,height=0.45\linewidth]{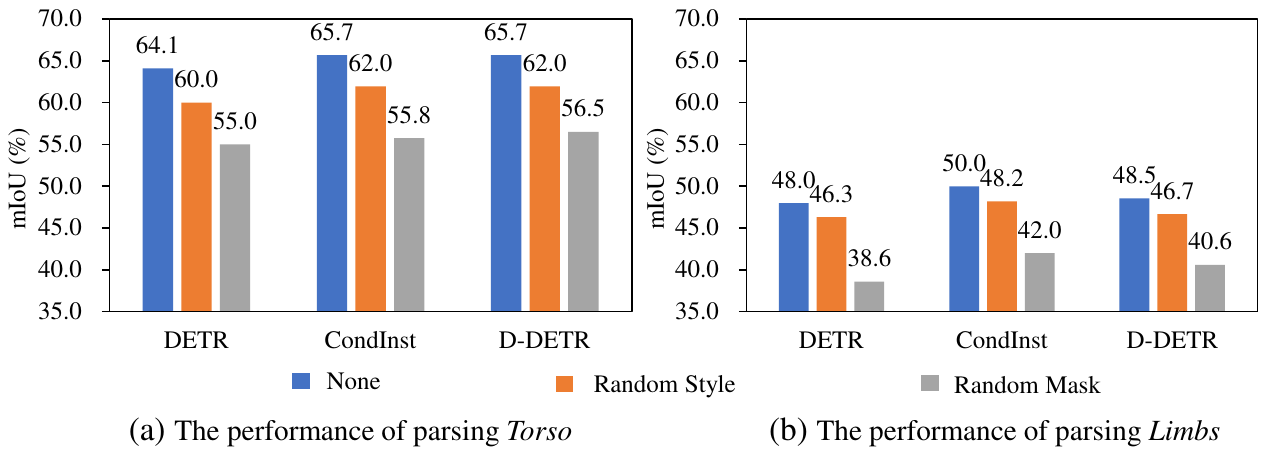}
\caption{The phenomenon of poor robustness: Existing parsing models are severely sensitive to external interventions, e.g., randomly changing image style or randomly masking partial body parts.}
\label{fig:context_change}
\vspace{-1.5em}
\end{figure}
Despite the great success, current methods are based on \textit{i.i.d} hypothesis, in which they intend to apply statistical models to acquire underlying associations between input images and labeled binary masks of body parts. 
However, such statistical models have shown to be vulnerable to visual variations.
For example, when applying state-of-the-art human parsers to images that are with unseen styles (Fig.~\ref{fig:domain_change}), the performance dramatically drops by nearly 20\%. Furthermore, when body parts are intervened by changing their surrounding contexts (Fig.~\ref{fig:context_change}), the parsing models intend to be severely sensitive to variations of the external environment.
Inspired by the Common Cause (CC) principle \cite{1956The}, these phenomenons can be analyzed via Structure Causal Model (SCM)~\cite{pearl2000causality,pearl2016causal,Judea2018thebookofwhy}. 
Specifically, associations between input variable and output variable are attributed to a confounder since it simultaneously influences the two variables (Fig.~\ref{fig:scm_analysis} (a)). Furthermore, the confounder can be viewed as a mix of causal and non-causal factors \cite{Lv_2022_CVPR,chen22iern}, where only causal factors decide both visual pattern of a person and corresponding semantic label. 
Therefore, as shown in Fig.~\ref{fig:scm_analysis} (b), the confounder is separated into non-causal factors that contain information with spurious correlations (e.g., \textit{``associations between image styles and body parts''}), and causal factors that include causal relations with relevant evidence (e.g., \textit{``shapes of body parts''}). 
In this way, given the confounder that is a mix of causal and non-causal factors, directly modeling associations from labeled data misleads parsing models to rely on non-causal factors \textit{w.r.t} spurious correlations, resulting in phenomenons of model degradation. 
To make a correct parsing result, a parsing model needs to exploit underlying causal factors and relies on causal relations \textit{w.r.t} relevant evidence. After all, it is the characteristic of a body part such as shape, instead of the style or background, that decides what a body part is. 

Motivated by above analysis, we propose a causality inspired parsing paradigm, termed \textit{CIParsing}. 
According to the Independent Causal Mechanism (ICM) principle \cite{PetersJanzingSchoelkopf17}, causal factors and non-causal factors should satisfy two causal properties for human parsing: 1) causal diversity and 2) causal invariance. A parsing model in proposed CIParsing is required to construct latent representations for human parsing and those representations are enforced to meet two basic causal properties.
To this end, we regard the part-related informations including the content of a part and corresponding body context as the causal factors, while the part-irrelevant informations such as style and background are regarded as non-causal factors. 
In CIParsing, a \textbf{\textit{factor separation}} pipeline is designed to extract latent representations of separated factors, and a \textbf{\textit{causality integration learning}} is utilized to guide latent representations of separated factors to satisfy the two causal properties. 
In this way, the human parser learns to rely on causal factors \textit{w.r.t} relevant evidence rather than non-causal factors \textit{w.r.t} spurious correlations, thus alleviating model degradation and yielding improved parsing ability.  
In summary, our work has the following contributions:

\begin{figure}[t]
\centering
\setlength{\abovecaptionskip}{3.pt}
\includegraphics[width=0.95\linewidth,height=0.37\linewidth]{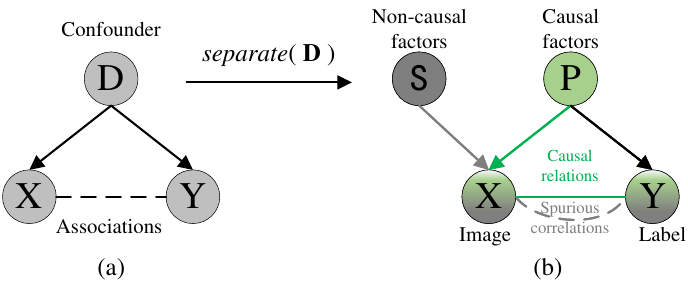}
\caption{Analysis of model degradation in MHP: \textbf{(a)} Underlying associations between variable X and variable Y are attributed to a confounder D that influences both X and Y. \textbf{(b)} The confounder is separated into causal and non-causal factors, where the latter ones bring spurious correlations.}
\label{fig:scm_analysis}
\vspace{-1.6em}
\end{figure}
\noindent\textbf{(1)} We propose a novel causality inspired human parsing paradigm termed \textit{CIParsing}, which particularly addresses model degradation problem in MHP via causality-based representation learning. 

\noindent\textbf{(2)} We formalize the MHP problem via a causal graph, and highlight two causal properties that causal factors of human parsing should satisfy. With the guidance of the causal graph, proposed CIParsing is designed in a plug-and-play fashion and can be integrated into any existing MHP model.

\noindent\textbf{(3)} Extensive experiments conducted on two challenging benchmarks (i.e., CIHP and MHP-V2) demonstrate the effectiveness and generalizability of the proposed method. Code and models are released for research purpose.

\section{Related Work}

\noindent\textbf{Multiple Human Parsing (MHP):}  Many works tackle the MHP problem by using either top-down or bottom-up methods. The series of top-down methods often obtain the bounding boxes of person instances. Then, the boxes are used to crop the image patches from the input image or image-level feature. Most top-down approaches assume each cropped patch includes only one person instance. Consequently, they are devoted to single-person parsing method by designing either isolated fully convolutional network \cite{swintrans,Liu_2022_CVPR,pose_SunXLWang2019} or regional CNN \cite{ins_seg:mask_rcnn,densepose:parsingrcnn,yang2020eccv}. For example, Mask RCNN~\cite{ins_seg:mask_rcnn} adopts Faster R-CNN \cite{obj_det:faster} to predict a bounding box for each person and extracts region-of-person from backbone network for performing single-person part segmentation. Following this idea, P-RCNN \cite{densepose:parsingrcnn} and RP-RCNN \cite{yang2020eccv} propose new variants of regional CNN respectively via contextual modeling and part re-scoring. In contrast, the bottom-up methods \cite{seg_part:cihp,Parsing:Graphonomy,parsing:mhpwild,seg_part:mhp,parsing:DSPF} regard multiple human parsing as a segment-then-grouping pipeline, in which they usually perform semantic part segmentation at first and then correspondingly assemble predicted binary masks to person instances. The main drawback of two-stage (top-down or bottom-up) methods is the slow inference and their efficiency is limited by the number of person instances within an image~\cite{Mao2021pose,Shi_2022_CVPR}.
Inspired by the success of the one-stage paradigm in object detection \cite{obj_det:detr,obj_det:deformdetr,tian2019fcos} and pose estimation \cite{Mao2021pose}, Chen \textit{et al.} \cite{RepParser} propose to utilize the representative body points to jointly solve instance separation and part segmentation, which brings the first one-stage multiple human parsing model.

Although these different approaches vary in parsing manner, they all share a characteristic: they focus on learning instance-specific representation by adopting statistical methods that model all dependencies between RGB images and labeled masks. Different from previous methods, our method unifies causality-based properties into the parsing pipeline, in which a parsing model is guided to excavate underlying causal cues w.r.t relevant evidence for human parsing.

\noindent\textbf{Causality-based Modeling:} 
To date, many works \cite{Tang_2020_CVPR,Lv_2022_CVPR,Wang_2022_CVPR,10.5555/3291125.3291161} try to enhance vision recognition models by studying causal inference. For example, recent works \cite{Wang_2022_CVPR, Lv_2022_CVPR} resort to causality to solve the out-of-distribution generalization problem in image classification task. In particular, they are devoted to either acquiring invariant causal mechanisms \cite{pmlr-v139-mahajan21b} or recovering causal features \cite{10.5555/3291125.3291161,ChangZYJ20} for image classification problem. However, how to explore intrinsic causality behind multiple human parsing, still remains an open question. As a supplement to them, our method can be viewed as an early attempt to explore causal mechanism for robust human parsing models.


\section{Causality Inspired Multiple Human Parsing}

\subsection{Human Parsing from a Causal View}
\label{sec:causal_view}
Estimating semantics behind pixels has its own reason, like Hegel's saying, \textit{``What exists is reasonable''}. Before presenting details of proposed method, we firstly give two principles summarized from prior researches \cite{1956The,PetersJanzingSchoelkopf17,causalLearning,Lv_2022_CVPR} for causal representation learning:

\noindent\textbf{Principle 1 (Common Cause Principle)~\cite{1956The}:} \textit{if two variables are statistically dependent, then there is a third variable that causally influences both of them and explains all the dependencies in the sense of making them independent}. It suggests an observed association between two variables may be resulted from other variables (e.g., confounder), and only causal variable decides the true essence of the association.  

\noindent\textbf{Principle 2 (Independent Causal Mechanism)~\cite{PetersJanzingSchoelkopf17}:} \textit{The conditional distribution of each variable given its causes (i.e., its mechanism) does not inform or influence the other conditional distributions.} In other words, when there exists many variables that causally influences identical variable, these causal variables are then independent of each other. 

\begin{figure}[t]
\centering
\setlength{\abovecaptionskip}{3.pt}
\includegraphics[width=0.75\linewidth]{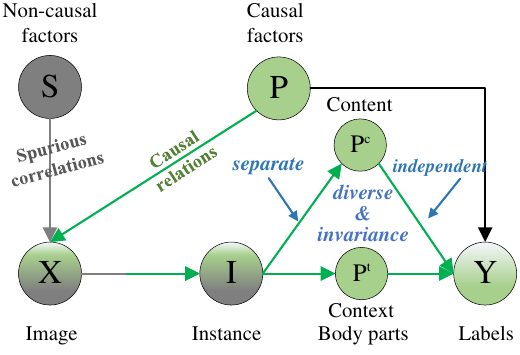}
\caption{The illustration of one-stage multiple human parsing via causal graph.}
\label{fig:cg}
\vspace{-1.5em}
\end{figure}
Based on the principles, we argue that the true factors of human parsing should satisfy two basic properties: 1) \textbf{\textit{Causal diversity}}, where causal factors independently decide the semantic label of each pixel and their explicit forms should be diverse. 2) \textbf{\textit{Causal invariance}}, where causal factors remain invariant to non-causal factors. However, complete factors are usually unobserved, which makes them hard to be directly reconstructed. As an alternative, we explicitly build part-relevant representations and guide them to satisfy the properties as the true causal factors. 
Inspired by that, we case study the one-stage MHP and formulate the pipeline in a detailed causal graph. As illustrated in Fig.~\ref{fig:cg}, the elements in the graph are defined as followed:

\noindent\textbf{Node $S$ \textnormal{and} $P$.} The former denotes non-causal factors that include part-irrelevant information with spurious correlations, while the latter denotes causal factors that contain characteristics of body parts.  

\noindent\textbf{Link $S \rightarrow X$, $P \rightarrow X$ \textnormal{and} $P \rightarrow Y$.} The first two links indicate the generation process of an image, where $S$ decides generation of instance-irrelevant information (e.g., image style or background) and $P$ decides the generation of body parts. The third link indicates the generation process of semantic labels.

\noindent\textbf{Node $X$ (Input Image \& Backbone Network).} A backbone network (e.g., ResNet-50) is used in this node. It usually outputs image-level feature maps $F$, taking RGB images with a mix of non-causal and causal factors as inputs.

\noindent\textbf{Link $X \rightarrow I$ (Instance Feature Extractor).} It firstly generates instance-aware convolutional kernels $K=\{k_i|i=1,...,n\}$ within one-stage framework. Then, the image-level feature maps are cast as instance-aware features $F_I$ using generated kernels, as formalized in Eq.~\ref{equ.inst_feat}. 
\begin{equation}
F_I = \text{\textit{Conv}}(F, K)
\label{equ.inst_feat}
\end{equation}

\noindent\textbf{Node $I$ (Instance-level Representation).} It contains characteristics of person instances and inherits a mix of causal cues and non-causal information from input images.

\noindent\textbf{Link $I \rightarrow P^t$ (Part's Context Extractor).} Contextual features of a part are separated from a corresponding instance feature. 

\noindent\textbf{Node $P^t$ (Part's Context).} Given a specific body part, this node contains characteristics of other body parts.

\noindent\textbf{Link $I \rightarrow P^c$ (Part's Content Extractor).} Content-based features of a part are separated from a corresponding instance feature. 

\noindent\textbf{Node $P^c$ (Part's Content).} It contains characteristics of a given part.

\noindent\textbf{Link $P^t \rightarrow Y$ (Part's Context Input for MHP).} This link indicates segmenting a given part using its body contexts.

\noindent\textbf{Link $P^c \rightarrow Y$ (Part's Content Input for MHP).} This link indicates segmenting a given part using its contents.

\noindent\textbf{Node $Y$ (Part Segmentation).} Pixel-wise classifiers used in this node predict binary masks for each part.
\begin{align}
M = \phi(F_P)
\label{equ.part_pred}
\end{align}
where $\phi(\cdot)$ is the standard softmax function and $M$ denotes the categorical binary masks with $C$ part classes. 

With the guidance of the causal graph (Fig.~\ref{fig:cg}), we elaborately introduce proposed CIParsing, which involves two major modules: Causal Factor Separation (CFS) and Causality Integration Learning (CIL). 

\subsection{Causal Factor Separation}
\begin{figure}[t]
\centering
\setlength{\abovecaptionskip}{3.pt}
\includegraphics[width=0.99\linewidth]{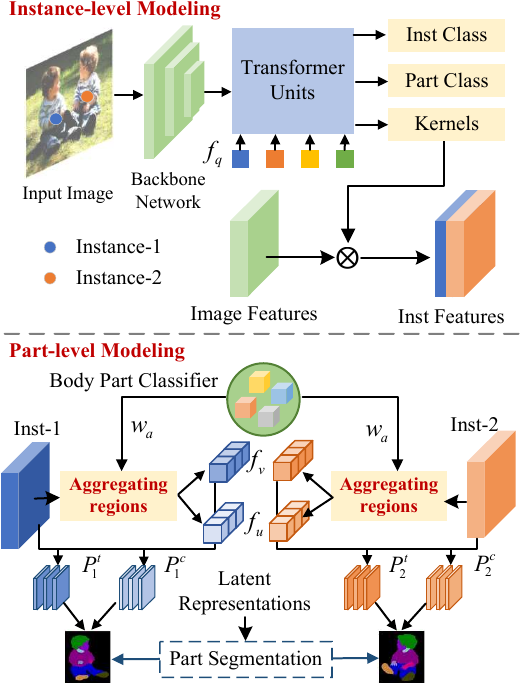}
\caption{The illustration of causal factor separation (CFS).}
\label{fig:framework}
\vspace{-1.5em}
\end{figure}

Following the causal graph, a multiple human parsing model is designed to construct latent representations with characteristics of body parts, thus facilitating accurate body part segmentation. As illustrated in Fig.~\ref{fig:framework}, the entire pipeline is divided into two following steps:

\noindent\textbf{(1) Instance-level Modeling:} As an initial step, a backbone network (e.g., ResNet-50) is used to map an input image into image-level feature representation $F \in \mathbb{R}^{d\times h \times w}$, where $d$ is the number of feature channels and $\{h, w\}$ denotes the spatial size. Following \cite{obj_det:detr,obj_det:deformdetr}, $N$ learnable embeddings $f_q \in \mathbb{R}^{N\times d}$, which represent a set of instance candidates, are used to detect person instances within an image. In specific, the learnable embeddings are decoded as instance-level features $f_o \in \mathbb{R}^{N\times d}$ through transformer-based computation units (e.g., DETR Transformer \cite{obj_det:detr} or D-DETR Transformer \cite{obj_det:deformdetr}), jointly taking image-level feature $F$ as input:
\begin{equation}
Input: \{F, f_q\} \xrightarrow{transformer} Output: \{f_o\}
\label{equ.inst_det}
\end{equation}

\noindent\textbf{(2) Part-level Modeling:} In general, taking instance-level feature $f_o$ as input, a person detector $\varPsi_{D}(\cdot)$ and a kernel generator $\varPsi_{K}(\cdot)$ are respectively used for person detection and part segmentation. Instead of directly applying generated instance-aware kernels to part segmentation, we firstly aggregate possible regions of body parts for extraction of causal factors.
Specifically, we adopt a part classifier $\varPsi_{A}(\cdot)$ to output part-level categorical distribution $c_a \in \mathbb{R}^{N\times C}$ for each instance. Notably, kernels $W_{a} \in\mathbb{R}^{d\times C}$ that come from part classifier $\varPsi_{A}(\cdot)$ contain global semantic information about part categories, since it needs to adapt to all instance-level representations learned from all samples.
Therefore, possible regions of body parts can be coarsely estimated by the affinity between the image-level feature and global semantic information, as illustrated in Eq.~\ref{equ.relation}:
\begin{equation}
\begin{array}{lll}
	A & \hspace{-0.5em}= Sigmoid((W_{a} \odot c_a) \otimes F ) &   \\
\end{array}
\label{equ.relation} 
\end{equation}
where $\odot$ is the broadcast element-wise multiplication and $\otimes$ denotes convolution operation. $A \in \mathbb{R}^{N \times C\times h \times w}$ is the estimated affinity field. For each body part, we aggregate features from corresponding regions, as illustrated in Eq.~\ref{equ.part_ct_ctx}:
\begin{equation}
\begin{array}{cll}
	f_v\{i\} & \hspace{-0.5em}= AdaPool(F \cdot A\{i\}) &   \\
	f_u\{i\} & \hspace{-0.5em}= AdaPool(F \cdot \sum\limits_{i\neq j}^{C}A\{j\}) &   \\
	s.t. &\hspace{-1em} i=1,2,..,C &
\end{array}
\label{equ.part_ct_ctx} 
\end{equation}
where $f_v \in \mathbb{R}^{N \times d}$ denotes part-level embeddings that focus on the contents of body parts. $f_u \in \mathbb{R}^{N \times d}$ represents part-level embeddings that focus on global contexts of body parts. The $AdaPool(\cdot)$ means the adaptive average pooling operation. Furthermore, we extend these two kinds of embeddings to spatial causal representations: 1) a spatial part representation $p^c \in \mathbb{R}^{N \times d \times h \times w}$ that stores causal cues based on the part's content, and 2) a spatial contextual representation $p^t \in \mathbb{R}^{N \times d \times h \times w}$ that includes causal cues from the part's contexts. This process is formalized as Eq.~\ref{equ.part_ctx_gen}:
\begin{equation}
\renewcommand\arraystretch{1.2}
\begin{array}{cc}
	F_{I}  = \varPsi_{K}(f_o) \otimes F    \\
	p^c  = \varPsi_{K}(f_v) \otimes F_I &  \\
	p^t  = \varPsi_{K}(f_u) \otimes F_I &   
\end{array}
\label{equ.part_ctx_gen} 
\end{equation}
where $F_I \in \mathbb{R}^{N \times d \times h \times w}$ denotes the instance-aware features and the kernel generator $\varPsi_{K}(\cdot)$ is a Multilayer Perceptron (MLP) function.

Given the separated representations (i.e., $p^c$ and $p^t$) that store intrinsic causal cues for part segmentation, we predict binary masks for a part either using the content of the part or its context, as formalized in Eq.~\ref{equ.sep_part_pred}.
\begin{equation}
\begin{array}{cll}
	M^c &\hspace{-0.5em} =&\hspace{-0.5em} \varPsi_{S}(p^c)    \\
	M^t &\hspace{-0.5em} =&\hspace{-0.5em} \varPsi_{S}(p^t)    
\end{array}
\label{equ.sep_part_pred} 
\end{equation}
where $M^c \in \mathbb{R}^{N \times C \times h \times w}$ and $M^t \in \mathbb{R}^{N \times C \times h \times w}$ are predicted binary masks. $\varPsi_{S}$ denotes a part segmentor with a linear convolution layer. 
\begin{figure}[t]
\centering
\setlength{\abovecaptionskip}{3.pt}
\includegraphics[width=0.99\linewidth]{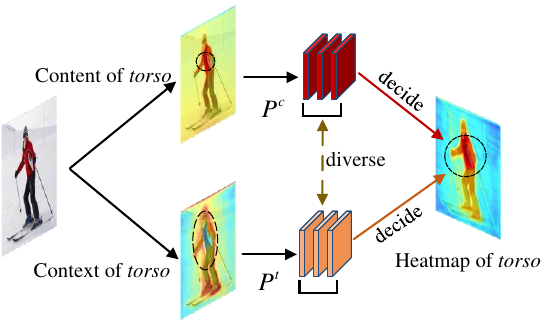}
\caption{The illustration of causal diversity: The heatmap of \textit{torso} is independently decided by two causal factors with different latent representations.}
\label{fig:causal_div}
\vspace{-1.6em}
\end{figure}
\subsection{Causality Integration Learning}
As discussed in Section \ref{sec:causal_view}, causal factors should satisfy two basic causal properties. In this section, latent representations of causal factors are guided to satisfy two properties, i.e., the causal diversity and the causal invariance. In particular, we define two following learning objectives.

\noindent\textbf{Learning Objective $\ell_{div}$:} As illustrated in Fig.~\ref{fig:causal_div}, the property of the diversity indicates that the intrinsic causal mechanism in a causal factor should be different from the mechanism of other causal factors. Since both content of a body part and relevant contextual body parts are regarded as causal factors, both spatial part representation and spatial contextual representation can be capable of part segmentation, but the explicit form of them should be diverse. Based on that, the learning objective for causal diversity is defined as Eq.~\ref{equ.part_seg}:  
\begin{equation}
\begin{array}{lll}
	\hspace{-0.5em}\ell_{part} &\hspace{-0.5em} = & \hspace{-0.5em} CE(M^c,G_{part}) + CE(M^t,G_{part})\\ [1ex]
	\hspace{-0.5em}\ell_{div} &\hspace{-0.5em} = &\hspace{-0.5em} T(\varPhi(p^c, G_{part}), \varPhi(p^t, G_{part})) + \ell_{part}
\end{array}
\label{equ.part_seg} 
\end{equation}
where $\varPhi(\cdot)$ is the aggregating function using the Eq.~\ref{equ.part_ct_ctx} with ground-truth binary masks $G_{part}$. $T(\cdot)$ denotes a similarity measurement based on cosine distance. $\text{\textit{CE}}(\cdot)$ denotes the cross-entropy loss.

\begin{figure}[t]
\centering
\setlength{\abovecaptionskip}{3.pt}
\includegraphics[width=0.75\linewidth]{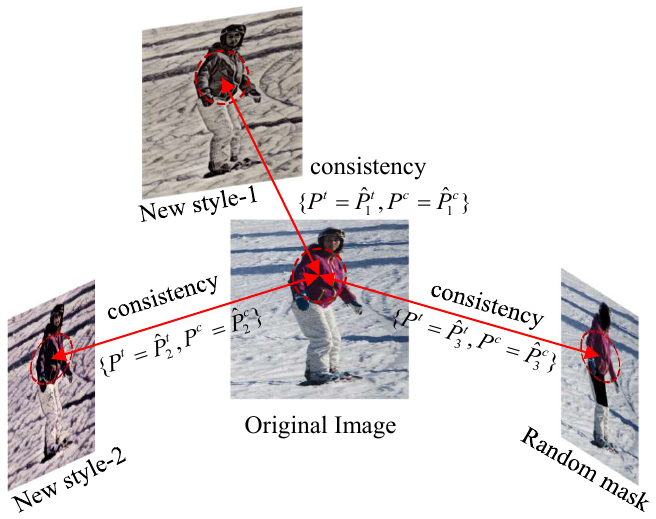}
\caption{The illustration of causal invariance: The representation of \textit{torso} is invariant to external changes (e.g., style changing or removal of partial body parts).}
\label{fig:causal_inv}
\vspace{-1.6em}
\end{figure}
\noindent\textbf{Learning Objective $\ell_{inv}$:} The property of causal invariance suggests that causal factors should remain invariant to the intervention of non-causal factors. In this way, when the non-causal factor is changed, the spatial part representation or spatial contextual representation should be semantically consistent. To this end, we simulate non-causal factors by generating fake images from a given image. Furthermore, semantics of each newly generated image remain consistent to the given one. As illustrated in Fig.~\ref{fig:causal_inv}, there exists $E$ newly generated images, which depict same visual content but involve different presentation styles. Latent representations of body parts, which are derived from the newly generated images, are denoted as $\{\hat{p^c_i} \in \mathbb{R}^{N \times d \times h \times w}, \hat{p^t_i} \in \mathbb{R}^{N \times d \times h \times w}\}_{i=1}^{E}$. Then, the learning objective for causal invariance is formalized through Eq.~\ref{equ.part_causal_inv}.
\begin{equation}
\renewcommand\arraystretch{1.2}
\begin{array}{cll}
	\hspace{-0.5em}\ell_{inv} &\hspace{-0.9em} = &\hspace{-0.9em} \frac{1}{E}\sum\limits_{i}^{E}|1 - T(\varPhi(p^c, G_{part}), \varPhi(\hat{p^c_i}, G_{part}))| + \\
	& & \frac{1}{E}\sum\limits_{i}^{E}|1-T(\varPhi(p^t, G_{part}), \varPhi(\hat{p^t_i}, G_{part}))|  
\end{array}
\label{equ.part_causal_inv} 
\end{equation}

To enable instance-level modeling and part-level modeling, the overall learning objective for optimizing a human parsing model is defined as Eq.~\ref{equ.model_learn}:
\begin{equation}
\begin{array}{cll}
	\ell_{det}&\hspace{-0.5em}=&\hspace{-0.5em}\text{\textit{CE}}(\delta_{cls}(F_I),G_{cls}) + \text{\textit{CE}}(c_a,G_{pcls}) + \\[1ex]
	& & \hspace{-0.5em}\text{\textit{SM}}(\delta_{reg}(F_I), G_{loc})\\ [1ex]
	\mathcal{L}&\hspace{-0.5em}=&\hspace{-0.5em}\ell_{det} + \ell_{div} + \ell_{inv}
\end{array}
\label{equ.model_learn} 
\end{equation}
where $\text{\textit{SM}}(\cdot)$ is the smooth-L1 loss and $G_*$ denotes corresponding ground-truth labels. $\delta_{cls}(\cdot)$ is the instance classifier and $\delta_{reg}(\cdot)$ is the coordinate regressor.
\section{Experiments}
\label{exp}
\subsection{Experimental Setup}
\noindent\textbf{Dataset and metrics:} We perform extensive experiments on two challenging benchmarks, i.e., CIHP~\cite{seg_part:cihp} and MHP-V2~\cite{seg_part:mhp}. The CIHP dataset contains 33k images in total, which are divided into two subsets for train/val (28k/5k images). The part-level annotations cover 19 part categories under various crowded scenes. The MHP-V2 is a commonly used dataset for instance-level human parsing, as it contains 20k images with 58 body part classes. Besides, it is split into 15k/5k images for train/val. Following official settings, we adopt \textbf{m}ean \textbf{I}ntersection-\textbf{O}ver-\textbf{U}nion (mIoU), \textbf{A}verage \textbf{P}recision based on \textbf{p}art (${\rm{AP}}^p$) and \textbf{P}ercentage of \textbf{C}orrectly parsed semantic \textbf{P}arts ($\rm{PCP}_{50}$) to evaluate multiple human parsing models. In particular, the ${\rm{AP}}^p_{vol}$ is the average of ${\rm{AP}}^p$ at IoU thresholds ranging from 0.1 to 0.9. 

\noindent\textbf{Implementation details:}
The proposed method is implemented based on Pytorch with eight NVIDIA Tesla V100 GPUs. Following common practice used in previous works \cite{densepose:parsingrcnn,yang2020eccv,obj_det:deformdetr}, the parsing models are trained using AdamW optimizer with a mini-batch size of 8. For the learning rate scheduler, the learning rate is initialized as 2e-4 and it is decreased by 10 at the 60-th for the 75-epoch setting. To simulate non-causal factors, we randomly change the style of a given image or randomly remove partial body parts. In particular, the pre-trained CLIPStyler~\cite{Kwon_2022_CVPR} is adopted to generate images with new styles, which include sketch style and cartoon style. During the training, each original image and its new form (either sketch style or cartoon style) are jointly utilized to guide the parsing model to learn causal invariance, and only original images contribute calculation of entire learning objective $\mathcal{L}$. 

\subsection{Comparisons with State-of-the-art Methods}
In this section, we compare the proposed method with state-of-the-art multiple human parsing methods on two evaluation datasets, i.e., CIHP and MHP-V2. In particular, we apply proposed CIParsing on three representative one-stage parsing models, including DETR~\cite{obj_det:detr}, condInst~\cite{inst_condinst} and D-DETR~\cite{obj_det:deformdetr}.

\noindent\textbf{CIHP:} Table~\ref{table:cihp} reports the comparison results against seven two-stage and three one-stage methods on CIHP \texttt{val}. From the results, we observe that our approach significantly outperforms existing one-stage models across all metrics. For example, in terms of the instance-level metric ${\rm{AP}}^p_{vol}$, the CIParsing based on D-DETR obtains a score of 58.4\% and outperforms D-DETR~\cite{obj_det:deformdetr} by 7.0\%. Similarly, the CIParsing respectively improves DETR~\cite{obj_det:detr} and condInst~\cite{inst_condinst} by 7.0\% and 8.1\%. Consistent improvements indicate the importance of causal modeling in one-stage human parsing methods. When comparing with two-stage approaches under the same settings, the CIParsing with D-DETR shows better performances than two-stage methods, such as M-RCNN~\cite{ins_seg:mask_rcnn} with 47.4\% ${\rm{AP}}^p_{vol}$, P-RCNN~\cite{densepose:parsingrcnn} with 53.9\% ${\rm{AP}}^p_{vol}$ and RP-RCNN~\cite{yang2020eccv} with 58.3\% ${\rm{AP}}^p_{vol}$. Built on a stronger backbone network, i.e., ConvNeXt-B, our method achieves the best performance across all parsing metrics. 
\begin{table}[t]
\centering
\setlength{\abovecaptionskip}{3.pt}
\renewcommand\arraystretch{1.}
\resizebox{0.99\linewidth}{!}{
	\begin{tabular}{c|c|cccc}
		\toprule
		Method       & \multicolumn{1}{c|}{Backbone}   
		& \multicolumn{1}{c|}{${\rm{AP}}^p_{vol}$} & \multicolumn{1}{c|}{${\rm{AP}}_{50}^p$} & \multicolumn{1}{c|}{mIoU} & ${\rm{PCP}}_{50}$\\ 
		\midrule
		\multicolumn{6}{l}{\textit{two-stage methods}} \\ 
		\midrule
		\multicolumn{1}{l|}{PGN~\cite{seg_part:cihp}}           & \multicolumn{1}{l|}{ResNet-101} 
		& 39.0 & 34.0   & 55.8   & 61.0      \\ 
		
		\multicolumn{1}{l|}{Graphonomy~\cite{Parsing:Graphonomy}}        & \multicolumn{1}{l|}{Xception}    
		& -    & -    & 58.6     & -         \\ \hline
		
		\multicolumn{1}{l|}{M-RCNN~\cite{ins_seg:mask_rcnn}}     & \multicolumn{1}{l|}{ResNet-50}  	 & 47.4 & 49.4  & 51.1    & 49.5     \\ 
		\multicolumn{1}{l|}{P-RCNN~\cite{densepose:parsingrcnn}}  & \multicolumn{1}{l|}{ResNet-50} 	 & 53.9 & 63.7 & 56.3    & 60.1   \\ 
		\multicolumn{1}{l|}{Unified~\cite{parsing:unified}}       & \multicolumn{1}{l|}{ResNet-101} 	 & 48.0 & 51.0  & 55.2   & -  \\ 
		\multicolumn{1}{l|}{M-CE2P~\cite{parsing:mce2p}}       & \multicolumn{1}{l|}{ResNet-101} 	 & - & -   & 59.5   & -  \\ 
		\multicolumn{1}{l|}{RP-RCNN~\cite{yang2020eccv}}  & \multicolumn{1}{l|}{ResNet-50}  	 & 58.3 & 71.6  & 58.2    & 62.2     \\ 
		
		\midrule	
		\multicolumn{6}{l}{\textit{one-stage methods}}\\ 
		\midrule
		\multicolumn{1}{l|}{DETR~\cite{obj_det:detr}} & \multicolumn{1}{l|}{ResNet-50}           & 48.2 & 48.1  & 53.3    & 50.5     \\ 
		\multicolumn{1}{l|}{condInst~\cite{inst_condinst}} & \multicolumn{1}{l|}{ResNet-50}    & 49.8 & 54.5  & 54.9  & 54.8   \\ 
		\multicolumn{1}{l|}{D-DETR~\cite{obj_det:deformdetr}} & \multicolumn{1}{l|}{ResNet-50}     & 51.4 & 56.2 & 54.4  & 55.5   \\
		\hline
		\multicolumn{1}{l|}{CIParsing (DETR)} & \multicolumn{1}{l|}{ResNet-50}     	 & 55.2 &  65.0 & 54.5 & 61.8   \\
		\multicolumn{1}{l|}{CIParsing (condInst)} & \multicolumn{1}{l|}{ResNet-50}     	 & 57.9 & 71.8 & 58.1 & 67.4  \\
		\multicolumn{1}{l|}{CIParsing (D-DETR)} & \multicolumn{1}{l|}{ResNet-50}     	 & 58.4 &   72.2 & 59.0  & 67.8   \\
		\hline
		\multicolumn{1}{l|}{CIParsing (D-DETR)} & \multicolumn{1}{l|}{ResNeXt-101}   	 & 59.6 &   74.6 & 61.5   & 69.3    \\
		\multicolumn{1}{l|}{CIParsing (D-DETR)} & \multicolumn{1}{l|}{ConvNeXt-B}   	 & 63.1 &  80.9 & 66.5   & 74.6  \\
		\hline
	\end{tabular}
}
\caption{Comparison with state-of-the-art methods on CIHP \texttt{val} set. }
\label{table:cihp}
\vspace{-1.5em}
\end{table}

\noindent\textbf{MHP-V2:} Table~\ref{table:mhpv2} presents comparisons with seven existing two-stage methods and three one-stage methods on MHP-V2 \texttt{val}. In line with the findings summarized from Table~\ref{table:cihp}, the proposed CIParsing significantly improves existing one-stage approaches by a large margin, leading to comparable or better parsing performances than that of two-stage methods. Based on this, one can conclude that our method performs general improvement on multiple human parsing and shows a good generalizability.

\subsection{Ablation Experiments}
In this section, we conduct extensive ablation studies on CIHP dataset. Specifically, we firstly investigate the effect of the core modules in CIParsing. Then, we particularly explore the generalizability and robustness of a parsing model that based on CIParsing. At the end of this section, we explore insights of the proposed method from a visualization perspective. We choose D-DETR with ResNet50 backbone as the baseline model and all parsing models are trained with 25 epochs throughout all ablative experiments.

\noindent\textbf{The effect of major modules.} To study the necessity of major modules in CIParsing, we start with a baseline model (i.e., D-DETR) which is built by instance-level modeling only. Then, we gradually improve the baseline model with: 1) a proposed causal factor separation (CFS) module that directly extracts causal factors via a basic learning objective (i.e., $\ell_{part}$ only); and 2) a causal integration learning (CIL) strategy that guides extracted latent representations of body parts to satisfy two causal properties (i.e., $\ell_{div}$ and $\ell_{inv}$). The experimental results listed in Table~\ref{table:ablation} indicate performance improvements after applying proposed techniques. In particular, three conclusions can be summarized: (1) Directly extracting latent representations of body parts is useful (e.g., ${\rm{AP}}^p_{vol}$ is increased from 51.4\% to 52.6\%), since it guides the parsing model to aggregate major characteristics of body parts (i.e., part's content and part's context) for part segmentation. (2) Compared with any single latent representation (i.e., either $p^c$ or $p^t$), utilizing all latent representations (i.e., $p^c$ and $p^t$) brings better predictive performances. (3) Causal integration learning is critical, as it further exploits the potential of a parsing model and achieves improvements of 3.9\% ${\rm{AP}}^p_{vol}$. These findings suggest that acquiring intrinsic causal mechanism is useful for multiple human parsing. 

\begin{table}[t]
\centering
\setlength{\abovecaptionskip}{3.pt}
\renewcommand\arraystretch{1.1}
\resizebox{0.99\linewidth}{!}{
	\begin{tabular}{|c|c|cccc}
		\hline
		\multicolumn{1}{c|}{Method}       & \multicolumn{1}{c|}{Backbone}      & \multicolumn{1}{c|}{${\rm{AP}}^p_{vol}$} & \multicolumn{1}{c|}{${\rm{AP}}_{50}^p$} & \multicolumn{1}{c|}{mIoU} & ${\rm{PCP}}_{50}$ \\ \hline
		\multicolumn{6}{l}{\textit{two-stage methods}}\\ \hline
		\multicolumn{1}{l|}{MH-Parser~\cite{parsing:mhpwild}}  & \multicolumn{1}{l|}{ResNet-101}   & 36.1    & 18.0  & -   & 27.0   \\ 
		\multicolumn{1}{l|}{NAN~\cite{seg_part:mhp}}        & \multicolumn{1}{c|}{-}       & 41.8   & 25.1  & - & 32.3   \\ 
		\multicolumn{1}{l|}{DSPF~\cite{parsing:DSPF}}        & \multicolumn{1}{l|}{ResNet-101}   & 44.3  & 39.0 & 41.4  & 42.3   \\ 
		\hline
		\multicolumn{1}{l|}{M-RCNN~\cite{ins_seg:mask_rcnn}}     & \multicolumn{1}{l|}{ResNet-50}   & 33.9 & 14.9   & -   & 25.1    \\	
		\multicolumn{1}{l|}{P-RCNN~\cite{densepose:parsingrcnn}}  & \multicolumn{1}{l|}{ResNet-50}  & 40.5 & 27.4   & 36.1   & 38.3     \\ 	
		\multicolumn{1}{l|}{M-CE2P~\cite{parsing:mce2p}}       & \multicolumn{1}{l|}{ResNet-101}    & 42.7 & 34.5 & 41.1  & 43.8          \\ 	
		\multicolumn{1}{l|}{RP-RCNN~\cite{yang2020eccv}}  & \multicolumn{1}{l|}{ResNet-50}   & 45.2 & 40.5  & 37.3    & 39.2        \\
		\hline
		\multicolumn{6}{l}{\textit{one-stage methods}}\\ \hline
		\multicolumn{1}{l|}{DETR~\cite{obj_det:detr}} & \multicolumn{1}{l|}{ResNet-50}    & 33.7 & 12.1 & 30.5  & 25.1       \\ 
		\multicolumn{1}{l|}{condInst~\cite{inst_condinst}} & \multicolumn{1}{l|}{ResNet-50}   & 36.5 & 18.7   & 26.2 & 30.1       \\ 
		\multicolumn{1}{l|}{D-DETR~\cite{obj_det:deformdetr}} & \multicolumn{1}{l|}{ResNet-50} & 36.6 & 17.2 & 35.7 & 32.4      \\ 
		\hline
		\multicolumn{1}{l|}{CIParsing (DETR)} & \multicolumn{1}{l|}{ResNet-50}      & 38.8 &  22.5 	& 23.9  & 34.7   \\
		\multicolumn{1}{l|}{CIParsing (condInst)} & \multicolumn{1}{l|}{ResNet-50}     	& 41.9 & 31.3 & 28.2   & 41.0  \\
		\multicolumn{1}{c|}{CIParsing (D-DETR)} & \multicolumn{1}{l|}{ResNet-50}    & 45.8 & 39.0  &  38.1 & 49.9     \\ 
		\hline
		\multicolumn{1}{c|}{CIParsing (D-DETR)} & \multicolumn{1}{l|}{ResNeXt-101}   & 46.3 & 40.9 &  39.3 & 52.0     \\ 
		\multicolumn{1}{c|}{CIParsing (D-DETR)} & \multicolumn{1}{l|}{ConvNeXt-B}  & 48.9 &  48.6 & 41.8 &  57.4     \\
		\hline
	\end{tabular}
}
\caption{Quantitative performance comparison on MHP-v2 \texttt{val} set.}
\label{table:mhpv2}
\vspace{-1.5em}
\end{table}
\begin{table}[ht]
\centering
\setlength{\abovecaptionskip}{3.pt}
\renewcommand\arraystretch{1.1}
\resizebox{0.9\linewidth}{!}{
	\begin{tabular}{c|cc|c|ccc}
		\hline
		\multirow{2}{*}{Baseline} &  \multicolumn{2}{c|}{CFS}    & \multirow{2}{*}{CIL} & \multirow{2}{*}{${\rm{AP}}^p_{vol}$}  & \multirow{2}{*}{mIoU}  & \multirow{2}{*}{$\rm{PCP}_{50}$} \\ \cline{2-3}
		& $p^c$ & $p^t$   &  &   & &  \\ 
		\hline
		\checkmark	&   \multicolumn{2}{c|}{-}  &   -     & 51.4  &   54.4                    & 55.5    \\ \hline
		\checkmark	& \checkmark &       & -  &    52.4   & 54.6  & 57.1  \\ 
		\checkmark&  & \checkmark       & -   & 52.2 &    53.8     &  56.7  \\ 
		\checkmark&  \checkmark &  \checkmark  & -& 52.6  &  54.4   & 57.3    \\ \hline
		\checkmark	& \checkmark&\checkmark & \checkmark & \textbf{56.5}  &  \textbf{55.1}   & \textbf{64.5}  \\ 
		\hline
\end{tabular}}
\caption{Ablation study of major components. CFS means proposed causal factor separation pipeline. CIL denotes proposed causal intervention learning.}
\label{table:ablation}
\vspace{-1em}
\end{table}

\begin{figure*}
\centering
\setlength{\abovecaptionskip}{3.pt}
\includegraphics[width=0.99\linewidth,height=0.26\linewidth]{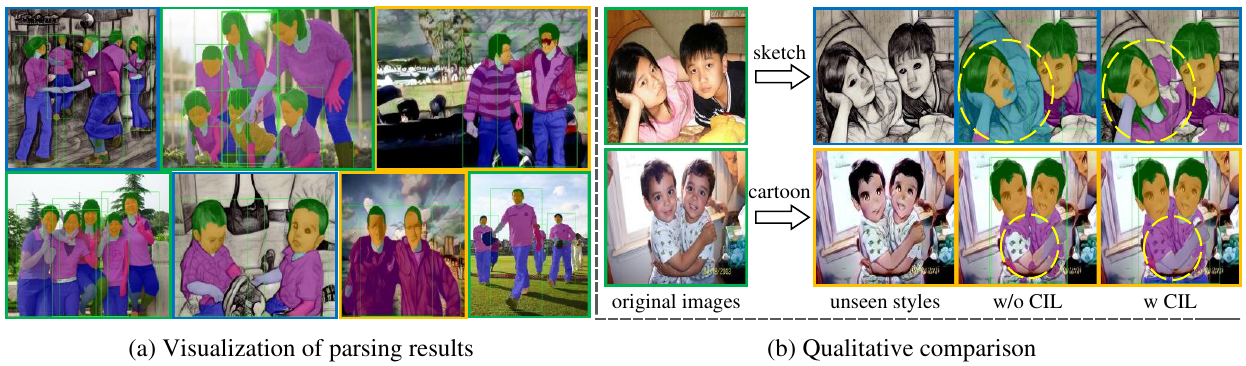}
\captionof{figure}{
	\textbf{Qualitative results:} (a) Visualization of parsing results with various image styles, including natural images (green box), cartoon images (orange box), and sketch images (blue box). (b) Qualitative comparison between a model trained without CIL and a model trained with CIL. Yellow dash circles indicate the detailed difference between them.
}
\label{fig:vis_results}
\vspace{-1.3em}
\end{figure*}

\noindent\textbf{Generalizability of a parsing model.} We investigate the generalizability of parsing models via a commonly used leave-one-domain-out protocol, where models are trained on images with a fixed style and tested on images with unseen styles. Specifically, two settings are used: 1) cartoon-style images are specified as unseen style; and 2) sketch-style images are set as unseen style. Summarized results listed in Table~\ref{table:dg} indicate that our approach surpasses all one-stage methods by a large margin at instance-level metric (i.e., ${\rm{AP}}^p_{vol}$ ). For example, when cartoon style is specified unseen domain (i.e., O-CIHP $\rightarrow$ C-CIHP), the CIParsing with D-DETR respectively outperforms condInst~\cite{inst_condinst}, RepParser~\cite{RepParser} and original D-DETR~\cite{obj_det:deformdetr} by 7.0\%, 4.2\% and 6.9\%. Consistent improvements are also observed when specifying sketch style as the unseen domain. 
This implies that CIParsing model is able to extract robust representations satisfying the property of causal invariance.
\begin{table}[t]
\centering
\setlength{\abovecaptionskip}{3.pt}
\renewcommand\arraystretch{1.3}
\resizebox{0.99\linewidth}{!}{
	\begin{tabular}{c|cc|cc}
		\hline
		\multirow{2}{*}{Method}      & \multicolumn{2}{c|}{O-CIHP $\rightarrow$ C-CIHP} & \multicolumn{2}{c}{O-CIHP $\rightarrow$ S-CIHP} \\ \cline{2-5}
		
		& \quad mIoU  &  ${\rm{AP}}^p_{vol}$   &\quad mIoU  &  ${\rm{AP}}^p_{vol}$ \\ \hline
		condInst~\cite{inst_condinst}    & \quad37.9  & 32.6& \quad 26.9  &  23.3 \\ 
		RepParser~\cite{RepParser}    & \quad38.2  & 35.4& \quad 26.1  &  24.7 \\ 
		D-DETR~\cite{obj_det:deformdetr}   & \quad37.9  & 32.7& \quad 26.3  &  23.0 \\ 
		\hline
		CIParsing (D-DETR)  &\quad \textbf{41.5}  & \textbf{39.6}  &\quad \textbf{30.4}  & \textbf{29.8} \\ 
		\hline	
\end{tabular}}
\caption{Investigating generalization capability of parsing models by testing them on unseen domain. Three types of image in CIHP are used: (1) original images (O-CIHP); (2) cartoon-style images (C-CIHP); and (3) sketch-style images (S-CIHP).}
\label{table:dg}
\vspace{-1.5em}
\end{table}

\noindent\textbf{Robustness of a parsing model.} To investigate the robustness of parsing models, we evaluate models on intervened images. Specifically, body parts of ``limbs'' in an image are randomly intervened by three ways: 1) ``Content only'', where only the content of ``limbs'' is preserved and other body parts are removed; 2) ``Random mask'', where body parts of ``non-limbs'' are randomly masked; and 3) ``Random style'', where the style of body parts are randomly changed. Each setting is repeated three times and then we report averaged evaluation results. Table~\ref{table:comp_ctx_adjust} summarizes quantitative comparisons under three predefined intervention settings. From the comparisons, we find that a CIParsing model without CIL achieves competitive results compared to the baseline model (see Fig.~\ref{fig:context_change} (b)). In sharp contrast, the robustness of a parsing model is significantly improved after using CIL, since the performances under two settings are separately improved by 4.7\%, and 7.7\%. This clearly indicates that modeling intrinsic causal cues is beneficial to model robustness.

\begin{table}[t]
\setlength{\abovecaptionskip}{3.pt}
\centering
\renewcommand\arraystretch{1.2}
\begin{subtable}[htb]{0.495\linewidth}
	\centering
	\resizebox{0.99\linewidth}{!}{
		\begin{tabular}{|c|c|c|}
			\hline
			\multicolumn{3}{|c|}{CIParsing (D-DETR) w/o CIL} \\ \hline
			\multirow{1}{*}{Target}               & Interventions            & mIoU  \\ \hline
			\multirow{3}{*}{Limbs}	& Content only               &   18.1\%  \\ 
			& Random Mask          &    41.4\%  \\ 
			& Random Style         &    47.5\%  \\ 
			\hline
		\end{tabular}
	}
	\caption{}
\end{subtable}
\begin{subtable}[htb]{0.495\linewidth}
	\centering
	\resizebox{0.99\linewidth}{!}{
		\begin{tabular}{|c|c|c|}
			\hline
			\multicolumn{3}{|c|}{CIParsing (D-DETR) w CIL}\\ \hline
			\multirow{1}{*}{Target}               & Interventions            & mIoU\\ \hline
			\multirow{3}{*}{Limbs} & Content only               &   26.4\% \\ 
			& Random Mask          &    45.3\% \\ 
			& Random Style         &    54.4\% \\ 
			\hline
		\end{tabular}
	}
	\caption{}
\end{subtable}
\caption{Quantitative comparison regarding model robustness.}
\label{table:comp_ctx_adjust}
\vspace{-1.5em}
\end{table}

\begin{figure*}[ht]
\centering
\setlength{\abovecaptionskip}{3.pt}
\includegraphics[width=0.9\linewidth]{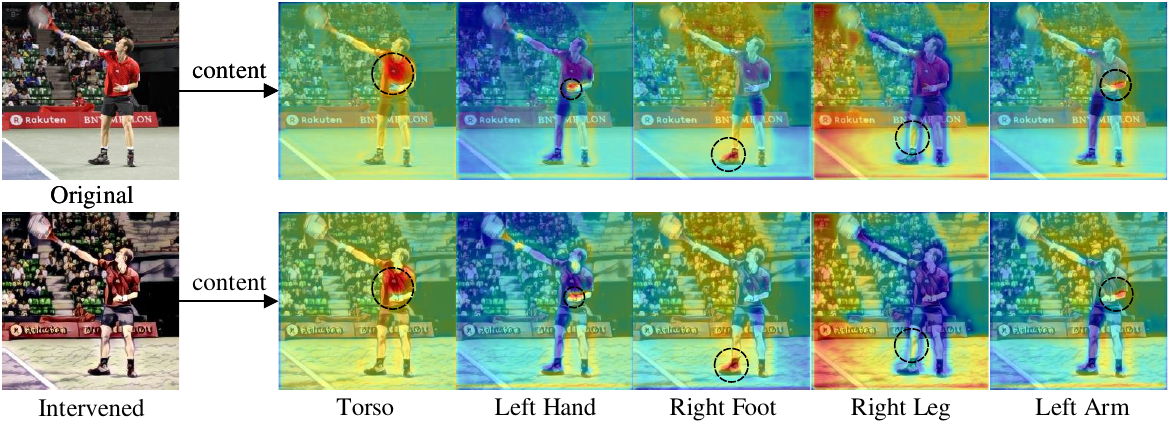}
\caption{Investigating the causal variance by visualizing latent representations with different style images.}
\label{fig:vis_inv}
\vspace{-1.3em}
\end{figure*}

\begin{figure}[ht]
\centering
\setlength{\abovecaptionskip}{3.pt}
\includegraphics[width=0.9\linewidth]{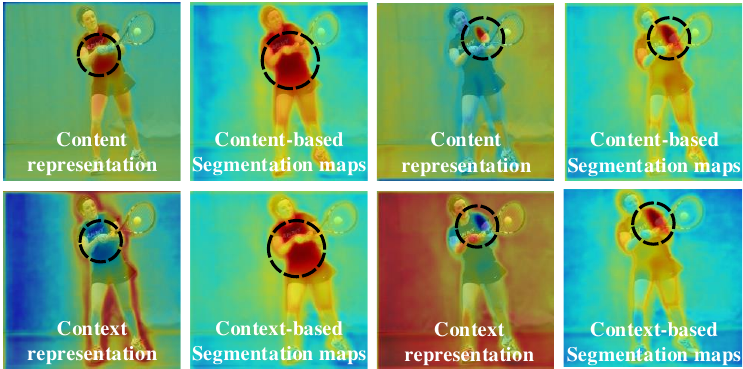}
\caption{Investigating the causal diversity by visualizing latent representations and corresponding segmentation maps. Circles indicate the target body part.}
\label{fig:vis_div}
\vspace{-1.5em}
\end{figure}

\noindent\textbf{Qualitative results.} We further investigate CIParsing models from a visualization perspective. As shown in Fig.~\ref{fig:vis_results} (a), CIParsing model can produce precise parsing results under various settings, such as crowd scenes (images in green boxes). When handling images with sketch style (images in blue boxes) or cartoon style (images in yellow boxes), corresponding parsing results remain good. To attain a further insight of learned model, two versions are compared: one is trained without CIL and another one is trained with CIL. The qualitative comparisons in Fig.~\ref{fig:vis_results} (b) shows that the one trained without CIL cannot well handle images with novel styles, but these failure cases can well be corrected by the one trained with CIL. Hence, it intuitively demonstrates that modeling causal cues is essential for making a correct parsing result, especially when input images are far different from the ones already learned by models (e.g., unseen styles). To investigate whether learned representations of causal factors are with the causal properties, we visualize latent representations that come from original image and intervened image, as illustrated in Fig.~\ref{fig:vis_inv}. From the visualization, we find that learned representations can well focus body parts regardless of the style of an image, indicating the invariance of learned latent representations. Furthermore, we compare two part segmentation maps that are respectively derived from two latent representations. As shown in Fig.~\ref{fig:vis_div}, we observe that identical segmentation maps can be predicted from two different latent representations, implying causal diversity of human parsing. For more details, we refer readers to supplementary materials. 

\section{Conclusion}
In this paper, we point out the insufficiency of one-stage multiple human models and present a new formulation of human parsing from a causal view. Then, we propose a causality-based one-stage human parsing method named CIParsing, aiming at modeling the intrinsic causal mechanism behind human parsing. Extensive experiments on two benchmarks prove the effectiveness and generalizability of our approach.

{\small
\bibliographystyle{ieee_fullname}
\bibliography{egbib}
}

\end{document}